\documentclass[11pt]{article}

\usepackage[margin=1in]{geometry}
\usepackage{amsmath,amssymb,booktabs,graphicx,hyperref,microtype,natbib,float}
\usepackage[tableposition=top]{caption}
\graphicspath{{../functional_sparse_operators/figures/}{figures/}}

\hypersetup{
  colorlinks=true,
  linkcolor=blue,
  citecolor=blue,
  urlcolor=blue,
  pdftitle={Selection, Representation, and Execution in Sparse Fourier Neural Operators},
  pdfauthor={Abdul Qadir Ibrahim and Martin Burger},
  pdfkeywords={Fourier neural operators, sparsity, compression, latency, negative results}
}
\title{\vspace{-1.2em}\textbf{
Selection, Representation, and Execution in Sparse Fourier Neural Operators}}
\author{%
Abdul Qadir Ibrahim \quad Martin Burger\\[0.4em]
\small Deutsches Elektronen-Synchrotron DESY, Notkestraße 85,
22607 Hamburg, Germany\\[0.3em]
\small \href{mailto:abdul.qadir.ibrahim@desy.de}{\texttt{abdul.qadir.ibrahim@desy.de}}
\quad
\href{mailto:martin.burger@desy.de}{\texttt{martin.burger@desy.de}}\\[0.2em]
\small \href{https://orcid.org/0000-0003-3452-8500}{ORCID: 0000-0003-3452-8500}
\quad
\href{https://orcid.org/0000-0003-2619-2912}{ORCID: 0000-0003-2619-2912}}
\date{Preprint --- \today}

\begin{document}
\maketitle

\begin{abstract}
Sparse representations are often expected to make models smaller and also reduce inference cost. For Fourier Neural Operators (FNOs), these objectives are not equivalent or do not always align: removing parts of the learned operator can leave the underlying transforms and dense computations unchanged, while changing the grid on which the model is evaluated can introduce overhead of its own. We therefore distinguish sparsity in the representation, in the stored parameters, in the theoretical operation count, and in measured runtime, and present an empirical study of several routes toward sparse FNOs that tests each transition between them separately. Coarsening the execution grid reduces the theoretical
cost without reducing measured latency, and adding a correction term recovers accuracy at the cost of making the model slower. Even an 83\% parameter reduction remains slower than the dense baseline under ordinary execution. These results motivate a stricter definition of useful sparsity: the deployed operator must preserve solution accuracy and map its reduced support to a genuinely cheaper execution path.
\end{abstract}

\section{Introduction}
\label{sec:intro}

Neural operators learn mappings between function spaces and provide a natural
setting for representations that are not tied to a fixed discretization
\citep{kovachki2023neuraloperator}. The Fourier neural operator (FNO) is a
popular example: its learned interactions are represented in Fourier space,
which makes spectral truncation, mode selection, and structured parameter
reduction natural compression strategies \citep{li2021fno}. Wavelet operators
provide a complementary representation that can capture localized scales and
interfaces \citep{tripura2023wno}. This motivates a simple question: can a
dictionary containing both Fourier and wavelet components be used to identify
which parts of an operator are actually needed?

We use such a Fourier--wavelet dictionary as a test-bed for support selection,
rather than as a proposed replacement for the FNO. The purpose is to study
whether different selection criteria identify removable operator groups, and
whether this type of sparsity leads to reductions in model size or computation.
A richer representation is therefore not assumed to be either more accurate or
more efficient.

This distinction matters because different notions of sparsity need not lead to
the same computational outcome. A model can be sparse in its basis
representation while still evaluating a full transform. It can contain fewer
stored parameters while relying on contractions that are slower than dense
matrix multiplication. Likewise, a lower analytical FLOP count does not need to result
in lower runtime once restriction, interpolation, synchronization, or
kernel-launch overhead is included. For FNOs, this issue is particularly
relevant because Fourier transforms, inverse transforms, and pointwise channel
mixing remain part of the computation even when only a subset of spectral
coefficients is retained.

We study these effects through a sequence of controlled compression experiments
for FNO-like operators. A sparsity mechanism is evaluated at three levels:
\begin{enumerate}
  \item \textbf{Selection:} whether operator groups can be removed without
  substantially degrading the learned solution map;
  \item \textbf{Representation:} whether this removal reduces stored
  parameters or an analytical operation count; and
  \item \textbf{Execution:} whether the resulting model is faster in measured
  hardware execution at comparable accuracy.
\end{enumerate}

Within a fixed Fourier--wavelet dictionary, we first compare support-selection
rules based on their effect on the solution loss, parameter magnitude, and
random removal. At tight budgets, loss-based selection gives lower mean error
than the parameter-norm and random controls on both heterogeneous Darcy flow
and one-dimensional Burgers flow. The result indicates that parameter magnitude
alone is not always a reliable measure of the importance of an operator group.

The mixed representation itself does not outperform a conventional FNO. We
therefore use the mixed model only to study selection within a heterogeneous
representation, while the conventional FNO remains the main architectural
reference throughout the paper. A spatial breakdown of the Darcy error refines
this comparison: the FNO's advantage is concentrated in the smooth bulk region,
while near permeability interfaces a functionally selected mixed support is
competitive with the independent FNO.

We also examine whether reductions obtained at the representation level carry
over to actual execution. Neither resolution coupling nor a large reduction in
stored parameters produces a faster model in our measurements, and a fine-scale
residual recovers accuracy at the cost of the intended saving. In each case the
reduction is real at the level it is claimed, but does not survive the
transition to measured execution.

Throughout these experiments, we observed the same issue repeatedly: nominal sparsity
does not guarantee computational sparsity. Reducing coefficients, parameters,
or analytical operations is useful only when the resulting structure can be
mapped to an execution path that is cheaper on the target hardware. This makes
the implementation of the sparse operator, including its transforms,
contractions, and kernels, part of the compression problem itself.

\paragraph{Contributions.}
The main contributions of this paper are:
\begin{enumerate}
  \item a distinction between basis sparsity, parameter sparsity, analytical
  compute reduction, and measured computational sparsity;
  \item a controlled comparison of loss-based, parameter-norm, and random
  support selection in a heterogeneous Fourier--wavelet operator;
  \item a comparison with a conventional FNO that separates the support
  selection question from the architectural comparison, including a regional
  breakdown showing that the FNO's advantage is concentrated in the bulk
  region rather than at permeability interfaces;
  \item two resolution-based compression experiments, including a fine-scale residual correction, evaluated with synchronized hardware measurements;
  \item an executable transformed-channel block prototype demonstrating that substantial parameter reduction need not reduce runtime without an
  appropriate sparse execution path; and
  \item a reproducible experimental protocol that reports both successful and unsuccessful compression strategies.
\end{enumerate}

\section{Background and related work}

In this section, we first review operator learning and the representation choices available to an FNO, then turn to sparsity and compression, where we introduce the four levels of sparsity used throughout the paper.

\subsection{Neural operators and representation choice}
An operator-learning model approximates a map such as
\begin{equation}
  \mathcal{G}^{\dagger}: a \longmapsto u,
\end{equation}
where $a$ may be a coefficient field or an initial condition and $u$ is the
corresponding PDE solution. Unlike a conventional finite-dimensional network,
the intended object is a map between function spaces; in practice, the map is
discretized on a grid. Neural operators and FNOs make this viewpoint explicit
\citep{kovachki2023neuraloperator,li2021fno}, while DeepONet provides another
important operator-learning construction \citep{lu2021deeponet}.

For an input field $v$, a schematic FNO layer is
\begin{equation}
  v \longmapsto \sigma\!\left(\mathcal{F}^{-1}\!\left[
  \widehat W(k)\,\mathcal{F}(v)(k)\right] + W_{\mathrm{loc}}v\right),
  \label{eq:fno-layer}
\end{equation}
where $\mathcal{F}$ is the spatial Fourier transform, $\widehat W(k)$ is a
learned channel-mixing matrix, and $W_{\mathrm{loc}}$ is a pointwise mixing
term. Mode truncation reduces the number of retained spectral coefficients,
but the complete FFT, inverse FFT, and pointwise paths may still execute.

Wavelet neural operators use localized multiscale representations and are
therefore a natural complement to Fourier operators, whose modes are global.
The wavelet literature provides the mathematical motivation for scale and
orientation subbands \citep{daubechies1992tenlectures,mallat1999wavelet}, while
the WNO formulation applies that structure to operator learning
\citep{tripura2023wno}. Adaptive-basis methods and mixtures of neural operators
provide related approaches to representation choice
\citep{zhao2026able,kratsios2024mono}.

\subsection{Sparsity and compression}

Classical parameter sparsity removes individual weights, filters, or blocks.
Lasso and group-lasso penalties are standard tools for shrinkage and group
selection \citep{tibshirani1996lasso,yuan2006grouplasso}, while pruning and
compression methods distinguish between the support identified during training
and the structure of the final exported network
\citep{han2016deepcompression,molchanov2017pruning}. Hard-thresholding and
greedy methods make this selection step explicit
\citep{blumensath2009iht,binev2011greedy}.

In this work, we distinguish four forms of sparsity:
\begin{description}
  \item[Basis sparsity:] fewer active Fourier shells or wavelet subbands.
  \item[Parameter sparsity:] fewer stored trainable scalars.
  \item[Analytical sparsity:] a lower theoretical FLOP count or operation
  proxy.
  \item[Computational sparsity:] lower measured memory use or wall-clock
  latency after inactive structure has been removed from execution.
\end{description}
A reduction at one level does not necessarily lead to a reduction at the next one. In the experiments below, we use these distinctions to separate support selection from reductions in model size and analytical cost, and from actual gains in execution.

Recent work on neural-operator compression shows why these distinctions matter.DINOZAUR replaces the dense FNO spectral multiplier with a diffusion-based parametrization that substantially reduces the parameter count and memory footprint. Its profiling also shows that factorized or gradient-augmented variants can remain slower because FFTs and additional tensor operations still account for much of the execution cost \citep{matveev2025dinozaur}. Operator boosting constructs stacked FNO, DeepONet, and CNO surrogates with 72--95\% fewer trainable parameters and mainly reports improvements in the accuracy--parameter trade-off \citep{shikhman2026boosting}. The spectrally-sparsified FNO (SS-FNO) augments each spectral layer with a lightweight diagonal gating selector whose implicit bias under stochastic gradient descent drives many frequency weights toward zero, and reports reduced active-mode counts together with lower memory footprint and computation cost
\citep{bacsa2025ssfno}. $\lambda$-FNO likewise uses a learned pruning matrix, and reports little change in inference time in its transfer-learning experiments \citep{xu2025lambdafno}.
A particularly close systems-level observation is reported for variable-spiking wavelet neural operators. Despite substantial spike sparsity, VS-WNO remains slower and more energy intensive than its dense WNO counterpart on a Jetson GPU because the deployed execution path continues to perform dense work \citep{yoo2026spikesparsity}.

These studies measure different aspects of compression. DINOZAUR includes
execution profiling, while the remaining methods report reductions in
parameters or active modes and, in some cases, aggregate runtime or memory
figures. Here we focus on a more specific question: for a fixed exported model, does the proposed reduction lower single-sample inference latency under ordinary execution? Parameter count, analytical cost, and measured runtime are related, but they are not equivalent.
More generally, hardware-aware sparsity results show that commodity dense
hardware benefits most from regular block or tile structure rather than
arbitrary zero patterns \citep{guo2020tilewise}. We therefore test directly whether the reductions considered here lead to measurable speedups for FNO-like operators.

\section{PDE benchmarks}

We use two standard benchmarks from the FNO literature, chosen so that the two representations in the mixed dictionary are tested against different kinds of structure. Darcy flow provides a two-dimensional problem with sharp coefficient contrasts and localized solution features, while Burgers flow provides a one-dimensional problem with nonlinear transport and gradients that evolve in time. Both use the public benchmark release of \citet{li2021fno}.

\subsection{Heterogeneous Darcy flow}

Darcy flow models steady flow through a porous medium:
\begin{equation}
  -\nabla\!\cdot\!\left(a(x)\nabla u(x)\right)=f(x),
  \qquad x\in(0,1)^2,
  \qquad u|_{\partial\Omega}=0.
  \label{eq:darcy}
\end{equation}
Here $a(x)>0$ is the heterogeneous permeability or diffusion coefficient,
$f$ is a forcing field, and $u$ is the resulting pressure-like solution. The
coefficient field is piecewise constant with a sharp contrast between two
values, so the solution is smooth over most of the domain but develops
localized gradients where the coefficient jumps. This makes Darcy a useful test
of whether a representation preserves both global structure and fine-scale
features. We use the data distribution and public benchmark release introduced
with the FNO study \citep{li2021fno}: fields are generated at high resolution
and subsampled to $141\times141$ for the experiments.
Figure~\ref{fig:qualitative} (top) shows a representative coefficient field
together with the corresponding solution.

\subsection{One-dimensional viscous Burgers flow}

\begin{figure}[H]
  \centering
  \includegraphics[width=0.92\linewidth]{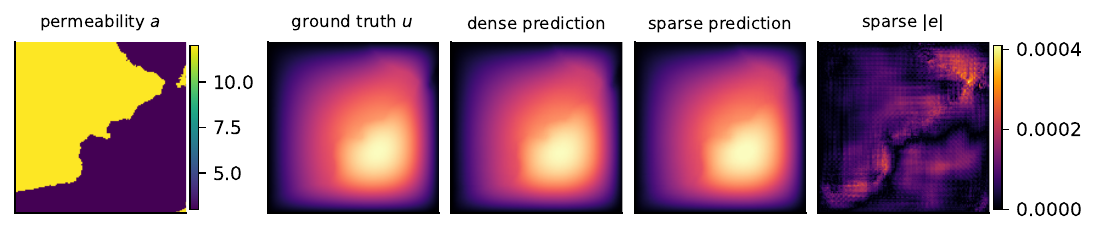}\par\vspace{0.6em}
  \includegraphics[width=0.92\linewidth]{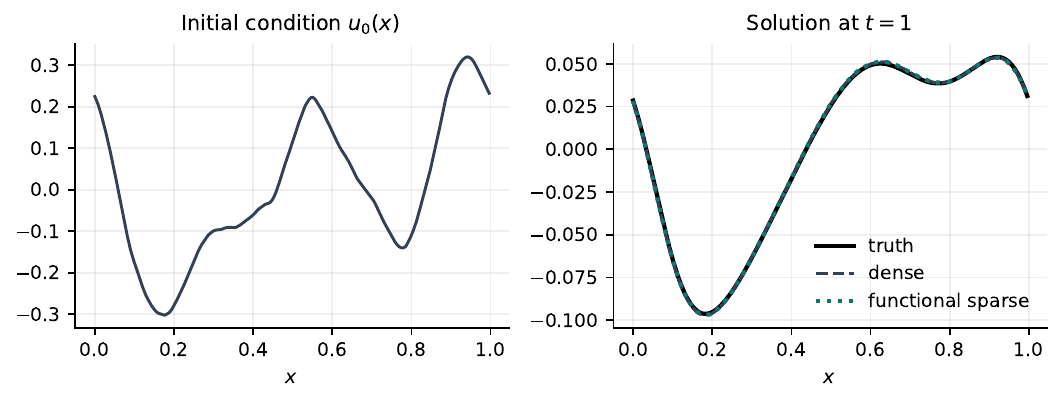}
  \caption{Representative held-out fields from the two primary benchmarks.
  Darcy (top) contains heterogeneous coefficient structure and localized
  solution features; Burgers (bottom) contains nonlinear transport and
  evolving gradients. The dense and sparse predictions shown for reference
  are from the mixed dictionary of Section~\ref{sec:methods} and its
  functionally selected support; both are visually indistinguishable from the
  ground truth at this scale.}
\label{fig:qualitative}
\end{figure}

The periodic one-dimensional Burgers equation is
\begin{equation}
  \partial_t u(x,t)+u(x,t)\,\partial_xu(x,t)
  =\nu\,\partial_{xx}u(x,t),
  \qquad x\in(0,1),\quad t\in(0,1],
  \label{eq:burgers}
\end{equation}
with a random periodic initial condition $u(x,0)=u_0(x)$. We use the FNO
benchmark release \citep{li2021fno} at $\nu=0.01$ and resolution 256. The
nonlinear transport term produces evolving gradients across different spatial
scales, making Burgers a complementary test to Darcy;
Figure~\ref{fig:qualitative} (bottom) shows a representative initial condition
and the solution at $t=1$.

\section{Methods} \label{sec:methods}

This section describes the models and procedures used in the experiments. We
first define the mixed Fourier--wavelet dictionary that serves as the test-bed
for support selection, together with the structured groups that act as the unit
of removal. We then give the functional selection rule and the parameter-norm
and random controls against which it is compared. The remaining subsections
describe the three execution routes: resolution coupling, a coarse model with a
fine-scale residual, and the transformed-channel block prototype.

\subsection{Mixed Fourier--wavelet operator dictionary}

The mixed model uses a lift--mix--project architecture. Each block contains a
Fourier expert, a wavelet expert, and group gates. The Fourier branch groups
global modes into shells, while the wavelet branch groups localized
coefficients into approximation and detail subbands. The operator contribution
is therefore expressed as a sum of structured operator components rather than
individual scalar weights. For Fourier groups $\mathcal{S}_j$ and wavelet groups $\mathcal{W}_r$, a
schematic block is
\begin{equation}
z_{\ell+1}=z_\ell+\mathcal{P}_{\ell}\left[
\sum_j g^{F}_{\ell,j}\Phi^{F}_{\ell,j}(z_\ell)
+\sum_r g^{W}_{\ell,r}\Phi^{W}_{\ell,r}(z_\ell)
\right],
\label{eq:mixed-block}
\end{equation}
where $\Phi^F$ and $\Phi^W$ denote the Fourier and wavelet contributions,
respectively, $g$ are group gates, and $\mathcal{P}_\ell$ denotes the local
mixing and projection operations. During dense training, the gates are scalar
parameters initialized and held fixed at one. During support selection, they
are set exactly to zero or one. Since the gates are not learned during dense
training, their values cannot be compensated by rescaling the corresponding
expert parameters.

For 2-D Darcy, shell $j$ contains retained frequencies satisfying
$\max(k_x,k_y)=j$. Each of the four blocks contains 12 Fourier shells. For each
of Haar, db2, and db4, a two-level 2-D wavelet decomposition contributes one
approximation group and three orientation groups (LH, HL, HH) at each level,
giving seven groups per wavelet family and block. The total number of
structured groups is therefore
\begin{equation}
N_{\mathrm{group}}=4\left[12+3(1+3\cdot2)\right]=132 .
\label{eq:ngroup-darcy}
\end{equation}

For 1-D Burgers, each block contains 16 individual Fourier modes. Each of the
three wavelet families contributes one approximation group and three
level-detail groups, giving
\begin{equation}
N_{\mathrm{group}}=4\left[16+3(1+3)\right]=112 .
\label{eq:ngroup-burgers}
\end{equation}
The wavelet transforms use periodization boundary handling, while the Fourier
transforms use orthonormal normalization.

\begin{figure}[H]
\centering
\begin{minipage}[b]{0.48\linewidth}
  \centering
  \includegraphics[width=\linewidth]{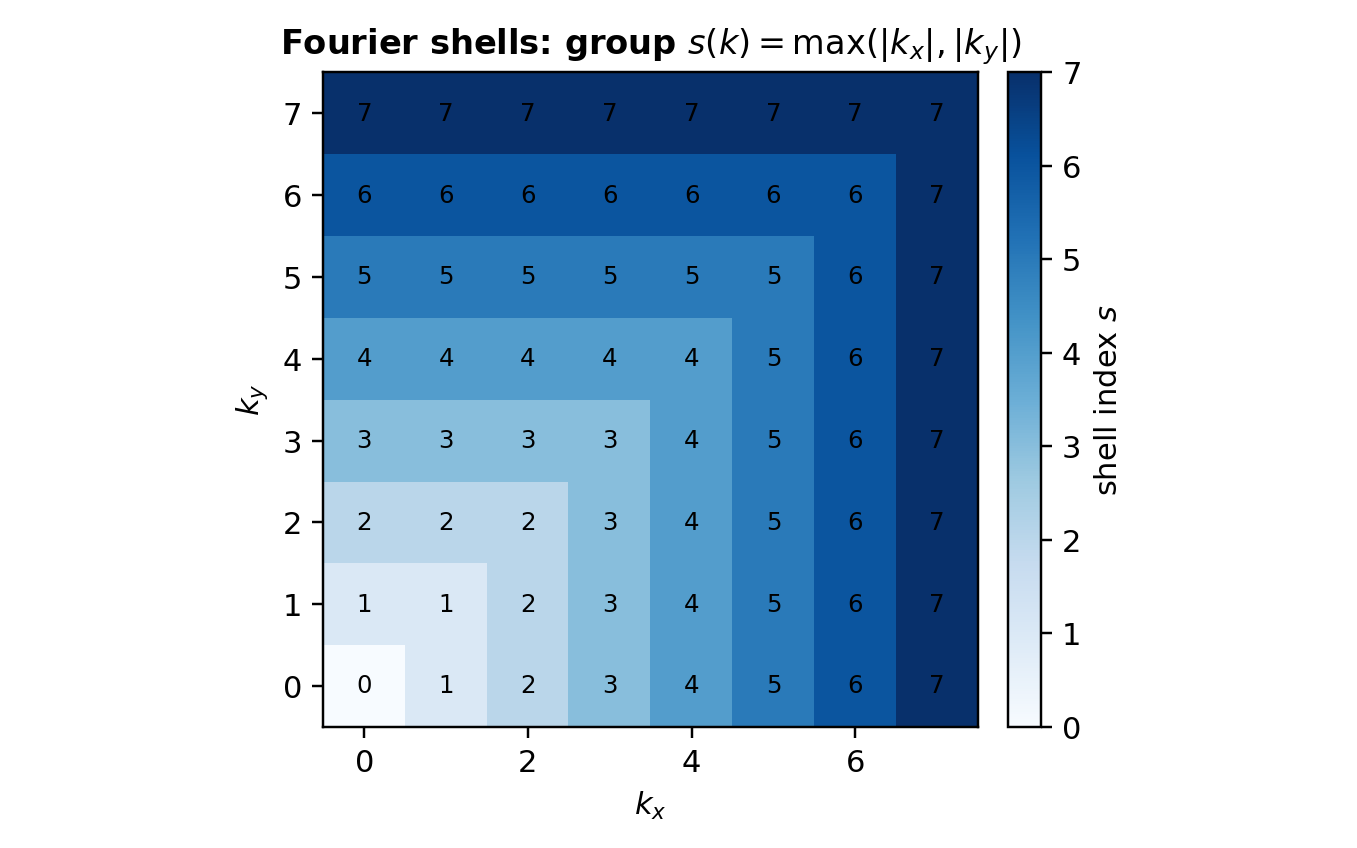}
\end{minipage}\hfill
\begin{minipage}[b]{0.48\linewidth}
  \centering
  \includegraphics[width=\linewidth]{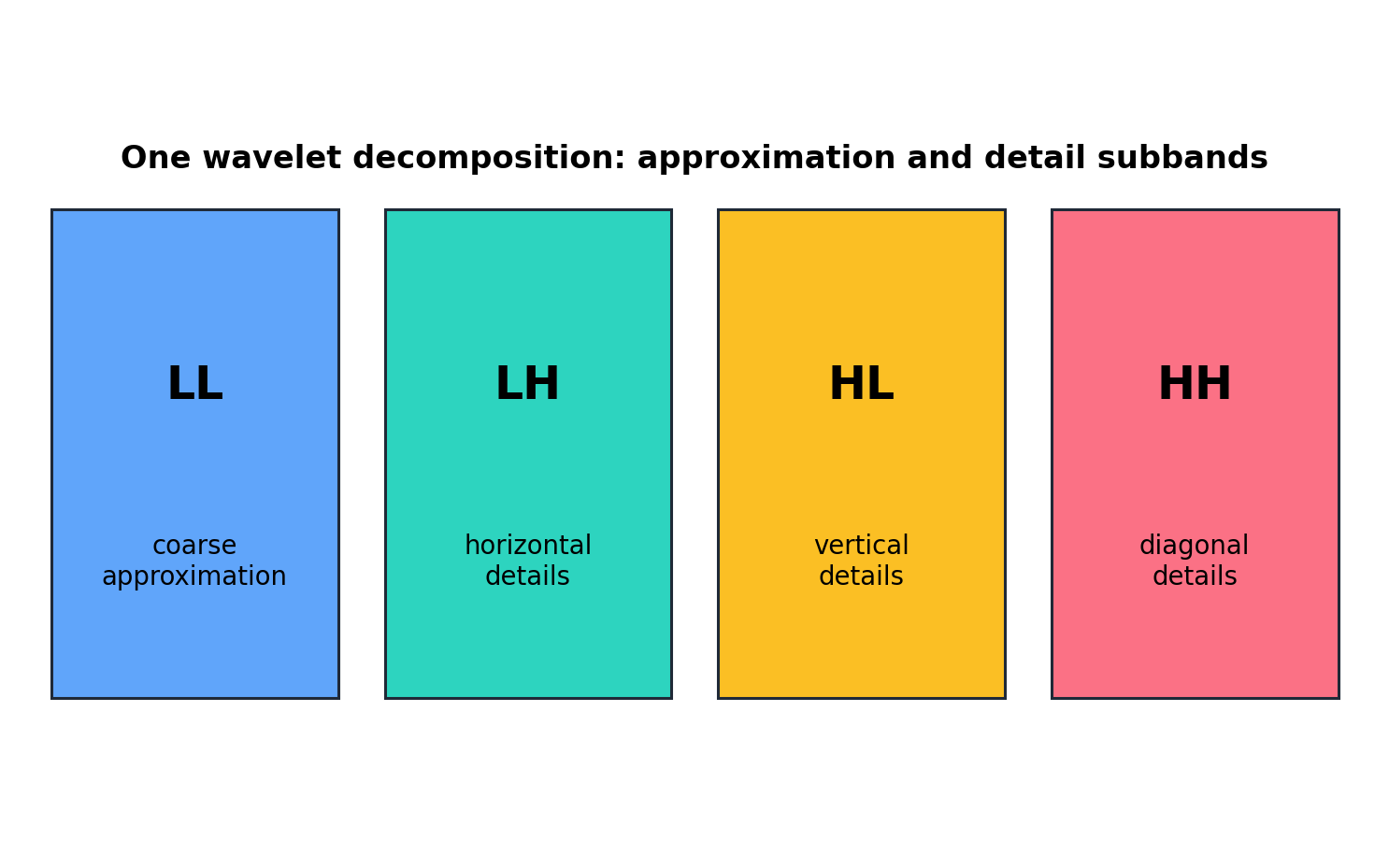}
\end{minipage}
\caption{The structured groups used by the mixed dictionary: global Fourier
shells (left) and localized wavelet subbands (right). Groups are the unit of
selection; an active group is either retained or removed as a complete operator
component.}
\label{fig:basis-groups}
\end{figure}

\subsection{Functional support selection}

Let $\mathcal{I}$ be the set of active groups and let
$\mathcal{L}_{\mathrm{cal}}$ denote the calibration loss. For each candidate
group $i$, we remove it from the current model and measure the change in loss,
\begin{equation}
\Delta_i(\mathcal{I})
=
\mathcal{L}_{\mathrm{cal}}(\mathcal{I}\setminus\{i\})
-
\mathcal{L}_{\mathrm{cal}}(\mathcal{I}).
\label{eq:ablation-effect}
\end{equation}
The greedy functional selector removes the group with the smallest effect and
recomputes the scores after each removal. Once the desired number of groups is
reached, the support is fixed and the model is retrained from the dense
checkpoint. This is an offline selection procedure, and its computational cost
is reported separately from the training cost.

The parameter-norm control does not rank the fixed gates. For group $i$, let
$\theta_i\in\mathbb{R}^{p_i}$ collect the trainable scalar parameters of the
corresponding expert component $\Phi_{\ell,j}^{F}$ or $\Phi_{\ell,r}^{W}$ in
Eq.~\eqref{eq:mixed-block}, counting real and imaginary parts separately for
the complex spectral weights. These are the parameters that are removed from
the model when the group is deactivated, and we refer to them as the
\emph{controlled} parameters of group $i$; parameters shared across groups,
such as the lifting, projection, and local mixing weights, are excluded. We
rank groups by the size-normalized score
\begin{equation}
m_i=\frac{\lVert\theta_i\rVert_2}{\sqrt{p_i}}.
\label{eq:parameter-norm-score}
\end{equation}
Groups with the largest scores are retained until the specified group-count or
parameter budget is reached. The normalization reduces the tendency to favor
groups simply because they contain more parameters, although group size is not
included directly in the ranking. As a result, a parameter-matched support can
contain fewer but larger groups than the support obtained by functional
selection. Parameter norms are also not invariant under general
reparameterizations within an expert. Random selection uses predeclared masks matched either by group count or controlled parameter count. Together, these controls separate the effect of support identity from group count, parameter count, and fixed-support
retraining.

\subsection{Resolution-coupled execution}

To test whether representational reductions translate into computational
savings, we couple the retained Fourier prefix $K_\ell$ at layer $\ell$ to a
latent grid of size $N_\ell$. The spectral, pointwise, and activation
operations are performed on this grid, with Fourier restriction before the
layer and interpolation afterward. This reduces the intended FFT and
pointwise workload, while the measured runtime also includes the cost of the
grid transitions and synchronization.

\subsection{Hybrid fine-scale residual}

Because coarsening can remove information needed to represent heterogeneous
coefficients, we also test a coarse branch with a lightweight fine-grid
residual:
\begin{equation}
u_{\mathrm{hybrid}}
=
u_{\mathrm{coarse}}
+
\rho\,u_{\mathrm{fine\ residual}}.
\label{eq:hybrid}
\end{equation}
The residual gate is a single global scalar
$\rho=\operatorname{sigmoid}(\alpha)\in(0,1)$, where the unconstrained logit
$\alpha$ is learned jointly with the model rather than fixed or annealed; it is
not per-channel or spatially varying. The logit is initialized to $-1$, so that
$\rho$ starts at $\operatorname{sigmoid}(-1)\approx0.269$. Although the
implementation supports an optional penalty on $\rho$, its coefficient is zero
in the reported experiment, so the gate is optimized only through the data
loss. This experiment tests whether the accuracy lost through coarsening can be
recovered without restoring the full computational cost.

\subsection{Executable transformed-channel basis}

The transformed-channel prototype acts on latent channels rather than spatial
frequencies. For a Fourier coefficient vector
$\widehat{v}(k)\in\mathbb{C}^C$, the spectral multiplier is written as
\begin{equation}
W(k)
=
B_{\mathrm{out}}^{-1}
S(k)
B_{\mathrm{in}},
\label{eq:fast-basis}
\end{equation}
where $S(k)$ contains explicit channel blocks. In the general construction,
$B_{\mathrm{in}}$ and $B_{\mathrm{out}}$ are separate real orthogonal
butterfly transforms acting on the $C$ latent channels, with
$B_{\mathrm{out}}^{-1}=B_{\mathrm{out}}^{\mathsf T}$. The spectral
coefficients and $S(k)$ are complex.

The phase-0 experiment fixes both transforms to the identity. We use $C=64$,
16 Fourier modes, and four layers, with both channel axes partitioned into
eight blocks of size 8. The dense-block model stores all
$8\times8=64$ channel-block pairs per layer, whereas the diagonal model stores
only the eight pairs $(q,q)$. Thus, ``diagonal'' refers to a block-diagonal
structure in channel space rather than diagonal matrices within each
$8\times8$ block. Inactive blocks are omitted from the module rather than
stored as zeros in a dense tensor. All three models retain the same channel
and mode dimensions.

The standard FNO, dense-block model, and diagonal model contain 549,569,
551,105, and 92,353 real scalar parameters, respectively. The diagonal
representation therefore reduces the parameter count relative to the standard
FNO by
\[
1-\frac{92{,}353}{549{,}569}=83.2\%.
\]
We test the identity basis first because learning an additional transform is
only useful if the corresponding sparse execution path is already competitive.
This separates the effect of the representation from limitations of its
execution.

\subsection{Experimental protocol}
\label{sec:protocol}

All canonical selection experiments use 1000 training, 200 validation, and 200
test samples. The mixed Darcy model has width 32, four blocks, 12 Fourier
shells, two wavelet levels, and 132 groups. The mixed Burgers model has width
32, four blocks, 16 Fourier modes, three wavelet levels, and 112 groups. Dense
models are trained first, after which the selected supports are fixed and
retrained from the corresponding dense checkpoint. The operating-point budgets were selected using one development seed and then held fixed. Results are reported over three seeds, with the remaining two serving as held-out confirmations.

\begin{table}[H]
\centering\scriptsize
\caption{Optimization and implementation details for the canonical accuracy
experiments. ``Best val.'' denotes the checkpoint with the lowest validation
relative $L^2$ error in physical units; the test split is evaluated once using
that checkpoint.}
\label{tab:protocol}
\resizebox{\linewidth}{!}{%
\begin{tabular}{llll}
\toprule
Item & Dense mixed dictionary & Fixed-support retraining & Conventional FNO \\
\midrule
Optimizer & AdamW & AdamW & AdamW \\
Learning rate / weight decay
& $10^{-3}$ / $10^{-4}$
& $10^{-4}$ / $10^{-4}$
& $10^{-3}$ / $10^{-4}$ \\
Schedule
& StepLR, 25 epochs, $\gamma=0.5$
& constant learning rate
& StepLR, 25 epochs, $\gamma=0.5$ \\
Epochs; batch (Darcy/Burgers)
& 100; 10/20
& 100; 10/20
& 100; 10/20 \\
Loss and checkpoint rule
& physical relative $L^2$; best val.
& physical relative $L^2$; best val.
& physical relative $L^2$; best val. \\
Initialization
& $1/C$ Gaussian spectral; final correction conv. $\mathcal N(0,(10^{-3})^2)$
& dense checkpoint
& $(C_{\rm in}C_{\rm out})^{-1}$ complex uniform spectral \\
Precision & FP32 & FP32 & FP32 \\
FFT normalization & orthonormal & orthonormal & PyTorch default (backward) \\
Wavelet boundary & periodization & periodization & not applicable \\
Gates & direct, fixed at 1 & exact binary mask, fixed & not applicable \\
\bottomrule
\end{tabular}}
\end{table}

For the conventional FNO comparison, the implementation concatenates spatial
coordinates, uses four spectral-plus-pointwise layers, retains both signed
vertical-frequency corners in two dimensions, and uses the canonical Darcy
padding. Widths were selected using a predeclared validation sweep. The primary
parameter-matched controls use width 32 for Darcy and width 56 for Burgers.

Latency is measured on one NVIDIA RTX A6000 using PyTorch 2.4.1 with CUDA 12.1,
FP32 tensors, eager execution, \texttt{model.eval()}, and
\texttt{torch.inference\_mode()}. We do not use \texttt{torch.compile} or CUDA
graphs. The benchmark code does not override the default TF32 settings.

For batch-1 resolution measurements, we use 20 warm-up iterations followed by
seven measurement blocks. Each block averages 100 forward passes for Burgers
and 50 for Darcy, with \texttt{torch.cuda.synchronize()} before and after each
block. We report the median of the seven block means together with the sample
standard deviation of those means. Analytical FLOP ratios are reported only as
proxies and are not interpreted as measured speedups.

\section{Results}

\subsection{Functional selection versus parameter norm and random support}

Table~\ref{tab:selection} summarizes the main within-dictionary selection
results. At the same group count, functional selection achieves lower mean test
error than parameter-norm selection on both Darcy and Burgers. The difference
is modest on Darcy but substantially larger on Burgers. Matching the controlled
parameter count gives a clearer separation on Darcy, where parameter-norm
selection retains fewer groups but produces considerably higher error. The
random controls show the same general pattern: support identity matters in
addition to the number of retained groups or parameters.

\begin{table}[H]
\centering\small
\caption{Canonical within-dictionary selection results. Test relative $L^2$
error is reported as mean $\pm$ SD. Dense, functional, and parameter-norm rows
use three runs; the group-matched random controls use six runs and the
parameter-matched random controls use four.}
\label{tab:selection}
\begin{tabular}{llccc}
\toprule
Benchmark & Method & Active groups & Controlled params. & Test error \\
\midrule
\multicolumn{5}{l}{\textit{Group-matched comparison}} \\
Darcy
& Dense & 132 & 2.733M & $0.013942\pm0.000268$ \\
& Functional & 79 & 1.307M & $\mathbf{0.012619\pm0.000283}$ \\
& Parameter norm & 79 & 1.835M & $0.013438\pm0.000720$ \\
& Random & 79 & 1.513--1.701M & $0.016986\pm0.002233$ \\
\addlinespace
Burgers
& Dense & 112 & 0.429M & $0.003143\pm0.000116$ \\
& Functional & 84 & 0.355M & $\mathbf{0.003132\pm0.000061}$ \\
& Parameter norm & 84 & 0.350M & $0.010046\pm0.003027$ \\
& Random & 84 & 0.313--0.334M & $0.004538\pm0.000654$ \\
\midrule
\multicolumn{5}{l}{\textit{Parameter-matched comparison}} \\
Darcy
& Functional & 79 & 1.307M & $\mathbf{0.012619\pm0.000283}$ \\
& Parameter norm & 50--52 & 1.306M & $0.051999\pm0.009393$ \\
& Random & 57--65 & 1.298--1.303M & $0.039454\pm0.010100$ \\
\addlinespace
Burgers
& Functional & 84 & 0.355M & $\mathbf{0.003132\pm0.000061}$ \\
& Parameter norm & 86 & 0.355M & $0.009083\pm0.001664$ \\
& Random & 92--96 & 0.351--0.359M & $0.003409\pm0.000253$ \\
\bottomrule
\end{tabular}
\end{table}

\begin{figure}[H]
  \centering
  \includegraphics[width=0.92\linewidth]{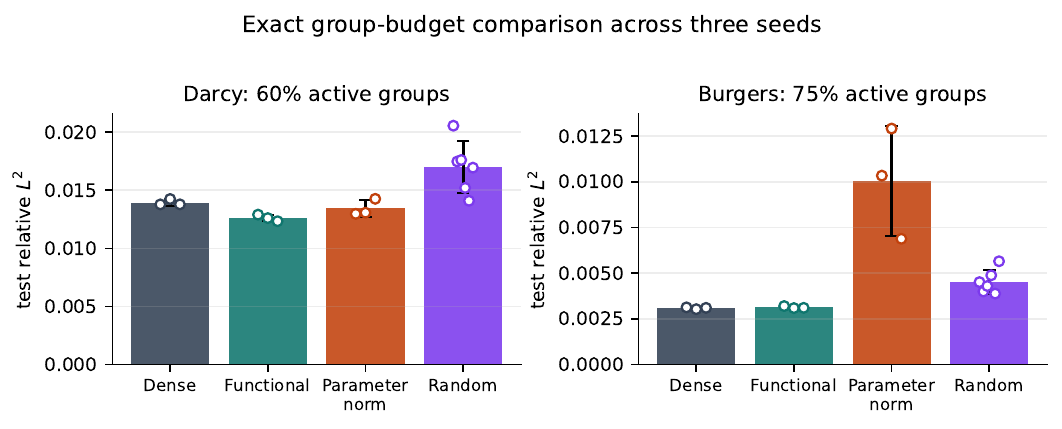}
  \caption{Group-matched comparison at 79/132 active groups for Darcy and
  84/112 for Burgers. ``Parameter norm'' refers to the size-normalized score in
  Eq.~\eqref{eq:parameter-norm-score}. These are the group-matched results from
  Table~\ref{tab:selection}. The functional result is a within-dictionary
  selection result rather than an architectural comparison with a conventional
  FNO.}
  \label{fig:selection}
\end{figure}

Figure~\ref{fig:selection} shows the group-matched comparison at the main
operating points. In Darcy, functional and parameter-norm selection are
relatively close in mean error, although the parameter-norm support uses
substantially more controlled parameters at the same group count. In Burgers,
the separation between the two selection rules is much larger.

Across the full exact-cardinality control, functional selection gives lower
test error in 34 of 36 comparisons with independently sampled random supports.
The operating-point budgets were chosen using the development run and then held
fixed for the confirmation runs. Additional support distributions and
retraining diagnostics are reported in Appendix~\ref{app:figures}.

Functional selection is an offline procedure and is more expensive than
parameter-norm ranking. The complete Darcy selection path takes approximately
22--25 minutes per run, while the corresponding Burgers path takes
approximately 4--5 minutes on the recorded GPU. These costs are incurred during
support discovery and are not included in inference latency. We therefore
interpret the advantage of functional selection as better support quality,
rather than lower overall training cost. Figure~\ref{fig:selection-dynamics} in Appendix~\ref{app:figures} provides
additional diagnostics. The retraining curves show the optimization behaviour
after fixing the selected support, while the random-support distributions show
how much the result varies with support identity.

\subsection{The mixed dictionary does not outperform a conventional FNO}

Table~\ref{tab:fno} compares the conventional FNO with the dense
mixed-dictionary model over three runs. The Darcy FNO is both smaller and more
accurate than the corresponding mixed model. The parameter-matched Burgers FNO
is also more accurate while using fewer parameters. These results show that the
mixed dictionary should not be interpreted as an architectural improvement over
a conventional FNO on either benchmark.

\begin{table}[H]
\centering\small
\caption{Comparison of the conventional FNO and the dense mixed-dictionary
model. Test relative $L^2$ error is reported as mean $\pm$ SD over three runs.
The mixed rows correspond to the pre-selection dense checkpoints.}
\label{tab:fno}
\begin{tabular}{llrrr}
\toprule
Benchmark & Model & Width & Parameters & Test error \\
\midrule
Darcy
& FNO & 32 & 2,368,001 & $\mathbf{0.012088\pm0.000042}$ \\
& Mixed dictionary & 32 & 2,742,209 & $0.014795\pm0.000252$ \\
\midrule
Burgers
& FNO & 56 & 421,769 & $\mathbf{0.002611\pm0.000103}$ \\
& Mixed dictionary & 32 & 437,953 & $0.004074\pm0.000106$ \\
\bottomrule
\end{tabular}
\end{table}

The mixed-model values in Table~\ref{tab:fno} correspond to the initial
pre-selection checkpoints, whereas the dense values in
Table~\ref{tab:selection} are obtained after an additional period of training.
They therefore represent different checkpoints of the same architecture and
should not be interpreted as repeated estimates of the same quantity.

\subsubsection{Where does the FNO's advantage come from?}

The global comparison in Table~\ref{tab:fno} does not say whether the
conventional FNO is uniformly more accurate or whether its advantage is
concentrated in a particular part of the domain. To check this for Darcy, we
split the test-set relative $L^2$ error into a smooth bulk region and a band
around permeability interfaces, using the same masked metric for the FNO and
for the mixed dictionary. Table~\ref{tab:darcy-regional} reports this split
for the independent FNO alongside the post-continuation dense checkpoint and
the 79/132 functional support from Table~\ref{tab:selection}; it therefore
uses a different mixed-model checkpoint than Table~\ref{tab:fno} for the
reasons given above.

\begin{table}[H]
\centering\small
\caption{Darcy interface/bulk error breakdown, using the checkpoints from
Table~\ref{tab:selection} (dense and 79/132 functional) and the independent
conventional FNO from Table~\ref{tab:fno}. Test relative $L^2$ error is mean
$\pm$ SD over three seeds. Bold marks the lowest error in each column.}
\label{tab:darcy-regional}
\begin{tabular}{lrrr}
\toprule
Model & Global error & Interface error & Bulk error \\
\midrule
Conventional FNO & $\mathbf{0.012088\pm0.000042}$ & $0.017402\pm0.000075$ & $\mathbf{0.010831\pm0.000026}$ \\
Dense mixed (132) & $0.013942\pm0.000268$ & $0.018667\pm0.000574$ & $0.012846\pm0.000204$ \\
Functional (79/132) & $0.012619\pm0.000283$ & $\mathbf{0.017045\pm0.000567}$ & $0.011599\pm0.000222$ \\
\bottomrule
\end{tabular}
\end{table}

The FNO's global advantage is driven almost entirely by the bulk region,
where its error is 10--16\% lower than either mixed-dictionary row. Near
interfaces the ordering changes: the 79/132 functional support reaches
slightly lower interface error than the independent FNO, and even the
untouched dense mixed model is within 7\% of it. Figure~\ref{fig:darcy-regional-errors}
in Appendix~\ref{app:figures} extends this comparison across the full
functional and parameter-norm budget sweep, with the FNO shown as a
reference band. One plausible explanation is that the wavelet-derived
families in the dictionary are suited to localized, non-smooth structure
(Section~\ref{sec:intro}), while the FNO's global Fourier basis spends its
representational budget on the smooth bulk region, where locality is not
needed.

The parameter-matched controls in Table~\ref{tab:selection} provide a separate
test of the support-selection result. Because the structured groups have
different parameter costs, matching the number of active groups does not
necessarily match the number of parameters. On Darcy, the parameter-matched
random and parameter-norm supports both produce substantially higher error than
functional selection. On Burgers, the parameter-matched random control is much
closer to the functional result, while parameter-norm selection remains worse.
We therefore report group-matched and parameter-matched comparisons separately.

The conclusion is limited but clear. Functional selection can identify useful
support within the heterogeneous dictionary, but this does not imply that the
dictionary itself is a better architecture. On the benchmarks considered here,
the conventional FNO provides the stronger dense baseline.

\subsection{Resolution coupling reduces a proxy, not measured latency}
On Burgers, the resolution-coupled $K=12,N=128$ model reduces the analytical
cost proxy to 50.9\% of the dense value, but its latency increases to
$1.474\pm0.033$~ms compared with $1.194\pm0.020$~ms for the dense model
(median $\pm$ repeat-block SD). A compact width-40 FNO achieves a validation
error of 0.002471 with a latency of 1.159~ms, reported as a point estimate.

The same pattern is more pronounced on Darcy. The $K=10,N=71$ model reduces
the analytical cost proxy to 27.2\% of the dense value, but its latency
increases to $2.070\pm0.023$~ms compared with $1.631\pm0.047$~ms for the dense
control. Its validation error also rises to 0.109249, compared with 0.011561
for the dense model, indicating substantial loss of solution accuracy. A
compact width-26 FNO achieves a validation error of 0.011557 at a latency of
1.623~ms. Because this latency is available only as a point estimate, the
0.008~ms difference from the dense median is not interpreted as a measured
speedup.

\begin{figure}[H]
  \centering
  \includegraphics[width=0.92\linewidth]{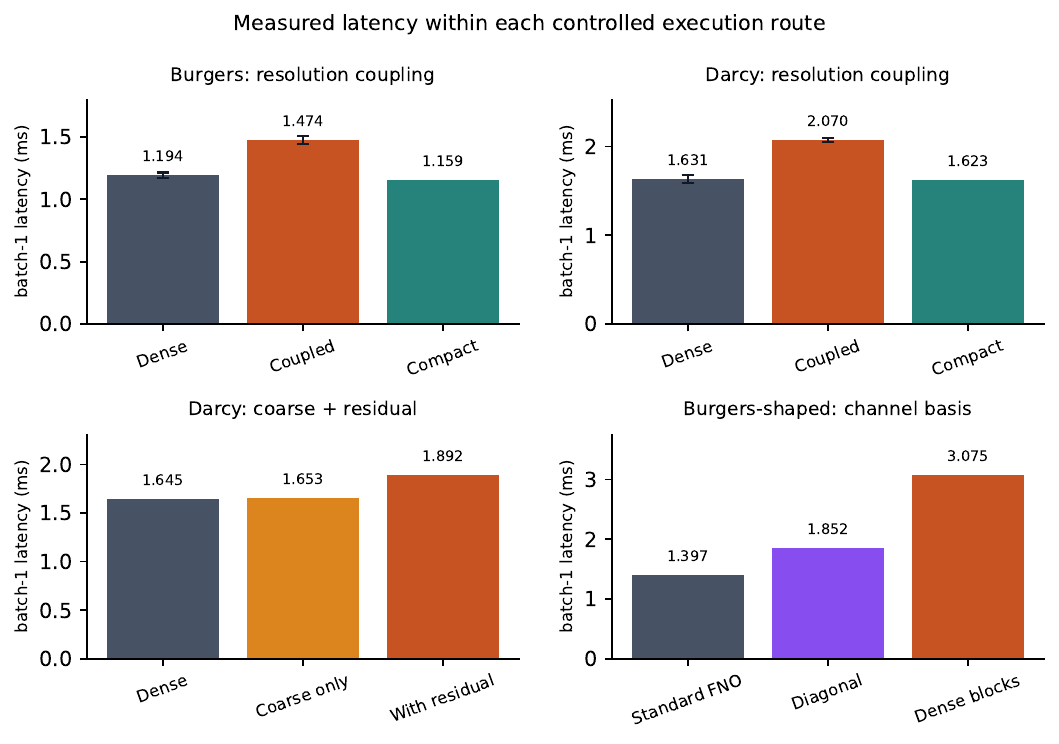}
  \caption{Batch-1 latency comparisons for four controlled execution routes.
  Each panel uses its own benchmark and implementation protocol, so absolute
  latency values should not be compared across panels. Error bars show the
  sample SD of seven repeat-block means where the raw repeats were retained;
  results without error bars are reported as point estimates. Resolution
  coupling and the residual route do not outperform their dense controls, and
  the transformed-channel implementations remain slower than the standard FNO.}
  \label{fig:negative-routes}
\end{figure}

Overall, reducing the execution grid lowers the analytical cost proxy but does
not reduce measured latency in these experiments. The additional grid
transitions offset the intended computational savings, while aggressive
coarsening can also remove information needed for the Darcy solution.
Figure~\ref{fig:negative-routes} compares this result with the other tested
execution routes.

\subsection{The fine-scale residual restores accuracy and cost}

The hybrid experiment makes the trade-off clear. Coarse-only execution has a
test error of 0.021322 and a latency of 1.653~ms. Adding the fine-scale
residual reduces the test error to 0.011888, close to the dense development
checkpoint's 0.012107, but increases latency to 1.892~ms. The corresponding
dense latency is 1.645~ms; the 0.008~ms difference from the coarse-only result
is treated as parity rather than a measured difference. At the reported
seed-42 checkpoint, the residual gate is 0.2677, essentially unchanged from its
initial value of approximately 0.269. Because this run uses no explicit gate
penalty, the near-constant value should not be interpreted as evidence of
learned gate selection. Instead, the test-error comparison shows that the
jointly trained fine-scale branch contributes while operating at approximately
its initialized scale. Thus, the residual restores accuracy but not the
intended computational saving.

Figure~\ref{fig:pareto} in Appendix~\ref{app:figures} therefore represents a
parameter--error trade-off rather than evidence of improved latency.

\subsection{Fast-basis parameters are not fast execution}

The fixed identity-basis phase-0 experiment provides a direct test of the
transformed-channel execution path. The diagonal representation reduces the
number of real scalar parameters by 83\%, but its batch-1 latency is
1.852~ms compared with 1.397~ms for the standard FNO. The dense-block
transformed representation is slower still, at 3.075~ms. We therefore do not proceed to learning the channel transform, since the fixed identity-basis implementation is already slower than the dense baseline. Adding a learned transform would also not address the underlying execution cost. Achieving a speedup would instead require a more efficient implementation, such as a fused
block-sparse kernel.

\section{Discussion}

\subsection{Summary of the compression routes}

\begin{table}[H]
\centering\small
\caption{Summary of the compression routes evaluated in this study.}
\label{tab:decision}
\scriptsize
\resizebox{\linewidth}{!}{%
\begin{tabular}{
p{0.23\linewidth}
p{0.32\linewidth}
p{0.35\linewidth}}
\toprule
Route & Main result & Limitation \\
\midrule
Functional group selection
& Preserves accuracy at tight budgets and outperforms parameter-norm selection within the mixed dictionary
& Does not by itself establish computational speedup \\

Parameter-norm selection
& Reduces the number of active groups and stored parameters
& Can remove groups that are important for solution accuracy \\

Resolution coupling
& Reduces the analytical cost proxy
& Grid transitions add overhead and coarsening can remove important fine-scale information \\

Coarse + fine residual
& Recovers much of the accuracy lost through coarsening
& The additional fine-scale computation removes the expected latency benefit \\

Transformed-channel blocks
& Reduces the stored parameter count by 83\%
& The tested execution path remains slower than the dense FNO and would require a more efficient sparse implementation \\
\bottomrule
\end{tabular}
}
\end{table}

Table~\ref{tab:decision} summarizes a common pattern across the experiments.
Selecting a reduced support can preserve solution accuracy, but this does not
necessarily produce a cheaper representation. Likewise, a cheaper
representation does not necessarily lead to faster execution. Computational
compression therefore requires all three steps: preserving the solution map,
reducing the representation or analytical cost, and realizing that reduction
through an efficient execution path.

\subsection{Interpreting the results}

Functional ablation can identify useful groups within a heterogeneous
dictionary, but this result is conditional on the chosen representation and
does not imply an architectural advantage over a conventional FNO. In our
experiments, the conventional FNO provides a stronger dense baseline. The main
result is therefore about support selection and execution rather than replacing
the FNO with a mixed Fourier--wavelet architecture. The regional breakdown in
Table~\ref{tab:darcy-regional} qualifies this conclusion: the FNO's advantage
is not uniform across the domain but concentrated in the smooth bulk region,
with the gap largely closing near permeability interfaces. A stronger global
baseline can therefore still leave room for a heterogeneous representation to
be locally competitive.

\subsection{From analytical cost to measured latency}

For an FNO layer on a grid of $N$ points, the nominal spectral cost is often
described by an $O(N\log N)$ transform together with channel mixing. Actual
latency also depends on memory movement, kernel fusion, batch size, transform
implementation, and whether inactive structures are physically removed from
execution. Hardware-aware sparsity studies similarly show that regular block or
tile structures are easier for dense hardware to exploit than arbitrary sparse
patterns \citep{guo2020tilewise,han2016deepcompression}. Our batch-1 measurements focus on single-sample inference, where kernel-launch
and transition overheads are visible. In the resolution-coupled experiment,
additional restriction and interpolation operations offset the reduction in
analytical cost. In the transformed-channel experiment, the sparse
representation replaces an optimized dense contraction with smaller block
operations that are slower under the tested implementation. In both cases, a
lower analytical cost does not translate into lower measured latency.
This is important when claiming computational sparsity. A zero
coefficient or masked parameter reduces the representation, but it reduces
execution cost only if the corresponding operation is also removed or replaced
by a cheaper implementation.

\subsection{Relation to adaptive-basis and mixture approaches}

Methods based on learned Fourier frames, adaptive bases, or mixtures of
operators are motivated by the fact that a single representation may not be
optimal for every problem \citep{zhao2026able,kratsios2024mono}. Our results
are consistent with this motivation, but they add an additional requirement:
the learned or selected structure must also admit an efficient implementation.
This helps explain why effective support selection within the mixed
dictionary can coexist with a stronger conventional FNO baseline. The first
concerns which components are useful within a fixed representation, while the
second concerns the choice of representation and architecture itself.

\subsection{Limitations and future work}

Several comparisons remain limited. The WNO experiment is restricted to a
Burgers diagnostic, while the Darcy run did not produce a valid final result.
We therefore make no matched comparison with a pure wavelet operator. Some
latency measurements also use development checkpoints and are reported
separately from the canonical accuracy results.

The parameter-norm baseline is more informative than raw gate magnitude: the
gates are fixed at one, and each group norm is normalized by the square root of
its parameter count. However, this score is not invariant to all possible
reparameterizations within a multi-layer expert. Our conclusions therefore
concern this particular norm-based control and do not imply that functional
selection outperforms every possible parameter-importance criterion.

Other limitations include the small number of independent runs, the limited
number of random-control trials, the use of a single NVIDIA A6000 GPU, and
ordinary PyTorch eager execution rather than a specialized sparse
implementation. The latency results therefore apply to the tested model sizes,
software stack, and hardware configuration. Different resolutions, hardware,
or sparse implementations may lead to different execution characteristics. The
interface/bulk breakdown in Table~\ref{tab:darcy-regional} is likewise based
on three seeds per model with no formal significance test, uses a fixed
band definition around permeability interfaces, and is reported only for
Darcy; we treat it as a descriptive regional trend rather than a general
claim that heterogeneous dictionaries are more accurate near interfaces.

The transformed-channel experiment is also limited by its current
implementation. A fused block-sparse kernel could change the latency result,
but this would require a separate implementation and profiling study. Larger
spatial grids or three-dimensional problems may likewise change the balance
between transform overhead and arithmetic cost.

Natural extensions of this work include:
\begin{enumerate}
  \item a resource-matched comparison with a pure wavelet operator;
  \item synchronized profiling of the canonical exported models; and
  \item evaluation of specialized sparse kernels if the resulting execution
        path can exploit the selected structure.
\end{enumerate}

\section{Conclusion}

We studied several approaches to reducing the representation and computational
cost of Fourier neural operators. Within a fixed heterogeneous
Fourier--wavelet dictionary, functional support selection preserves accuracy
and outperforms size-normalized parameter-norm selection at the tested
operating points. However, the mixed dictionary does not outperform the
conventional FNO baseline overall. Moreover, neither resolution-based reduction nor an
83\% reduction in stored parameters leads to lower measured latency under the
tested execution paths. A fine-scale residual recovers much of the accuracy
lost through coarsening, but also removes the expected computational saving.

These results show that parameter reduction and analytical cost reduction are
not sufficient measures of computational sparsity. A useful sparse
neural-operator representation must preserve the solution map and translate
its reduced structure into a cheaper execution path on the target hardware.
This requires considering the operator representation and its implementation
together, including transforms, memory movement, and the kernels used to
execute the retained structure.

\section*{Data and code availability}

Code is available at \url{https://github.com/aqibrahimbt/sparsity_is_not_speed}.
The version corresponding to this manuscript is archived under the tag
\texttt{arxiv-v1}. Large datasets, model checkpoints, and raw training outputs
are not included in the repository; the corresponding experimental protocols
and artifact names are documented there.

\section*{Acknowledgments}

This work was carried out at DESY using the Maxwell GPU computing
infrastructure. The authors acknowledge DESY for providing the computational
resources used in the reported experiments.

\bibliographystyle{unsrtnat}
\bibliography{references}

\appendix
\section{Supporting figures}
\label{app:figures}

The figures in this appendix provide additional diagnostics for the selection
and compression experiments reported in the main text. They are included here
to keep the main results focused on the primary accuracy and latency
comparisons.

\begin{figure}[H]
  \centering
  \includegraphics[width=0.92\linewidth]{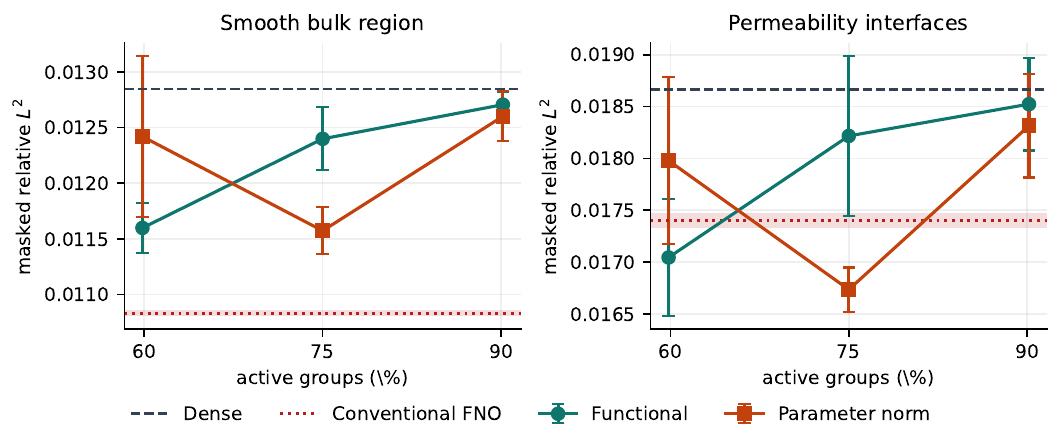}
  \caption{Regional error diagnostics for Darcy. The interface and bulk
  regions show how the remaining error is distributed spatially and provide
  information that is not visible from the global relative $L^2$ error alone.
  The dotted red line and shaded band show the mean $\pm$ SD of the
  independent conventional FNO (width 32, seeds 42--44, Table~\ref{tab:fno})
  under the same masked metric. The FNO is well below the mixed-dictionary
  curves in the bulk panel but sits inside their range in the interface
  panel, matching the summary in Table~\ref{tab:darcy-regional}.}
  \label{fig:darcy-regional-errors}
\end{figure}

\begin{figure}[H]
  \centering
  \includegraphics[width=0.76\linewidth]{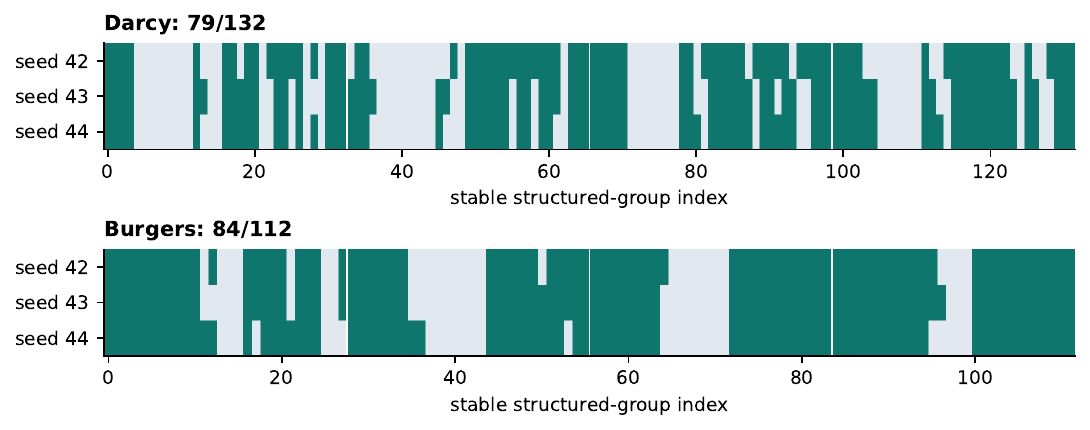}\par\vspace{0.5em}
  \includegraphics[width=0.76\linewidth]{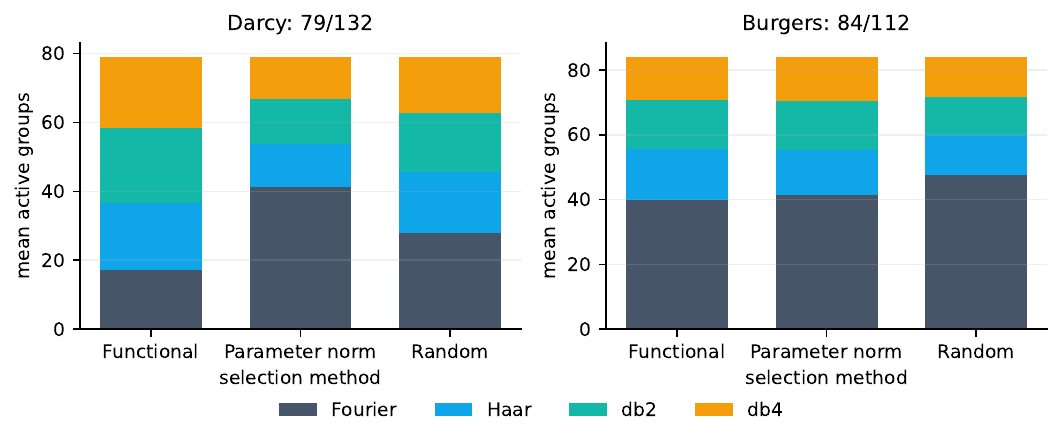}
  \caption{Diagnostics of the selected support. The top panel shows agreement
  between supports obtained in different runs, while the bottom panel shows
  the distribution of retained groups across Fourier and wavelet families.
  The comparison is made at the level of structured groups rather than
  individual scalar parameters.}
  \label{fig:support-diagnostics}
\end{figure}

\begin{figure}[H]
  \centering
  \includegraphics[width=0.76\linewidth]{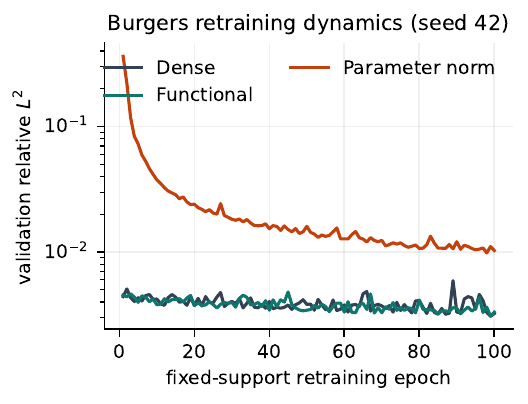}\par\vspace{0.5em}
  \includegraphics[width=0.76\linewidth]{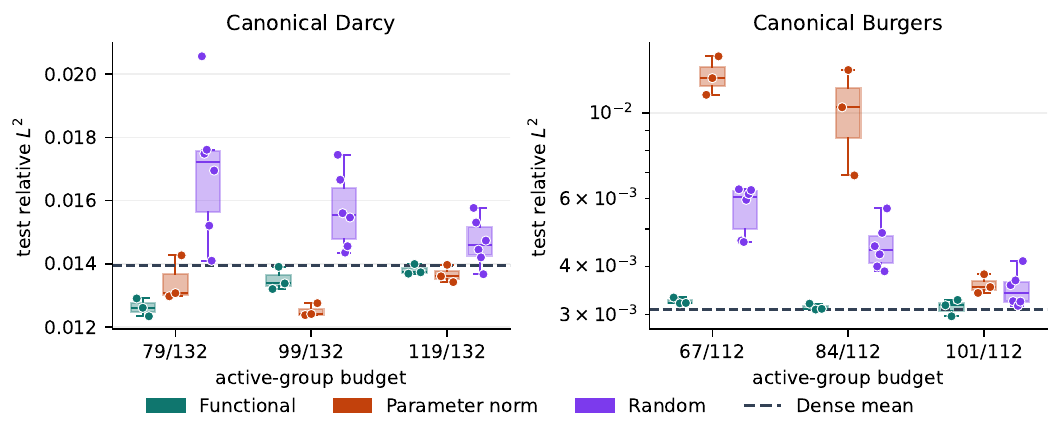}
  \caption{Additional checks on the support-selection experiment. The top
  panel shows retraining after the support is fixed, while the bottom panel
  shows the distribution of outcomes for the random-support controls.}
  \label{fig:selection-dynamics}
\end{figure}

\begin{figure}[H]
  \centering
  \includegraphics[width=0.92\linewidth]{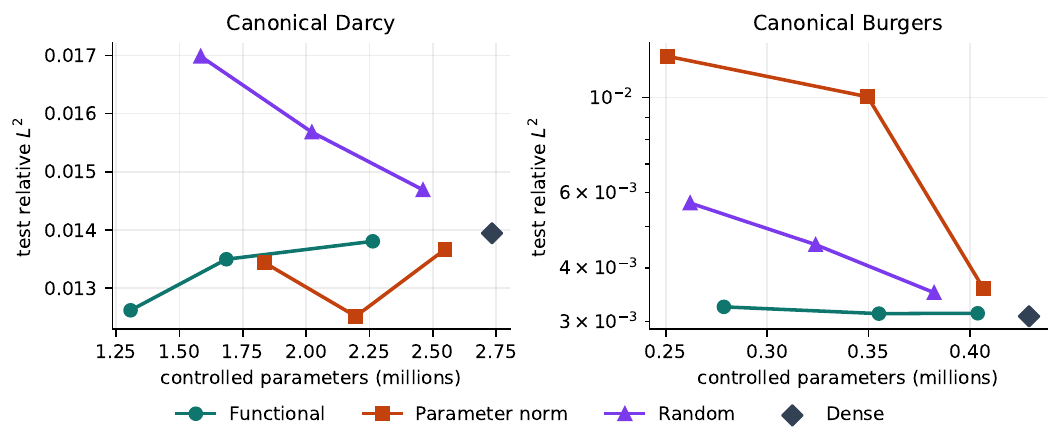}
  \caption{Parameter--error trade-offs for functional selection. The figure
  shows that parameter reduction can coexist with preserved accuracy, but does
  not by itself imply lower inference latency.}
  \label{fig:pareto}
\end{figure}

\begin{figure}[H]
  \centering
  \includegraphics[width=0.92\linewidth]{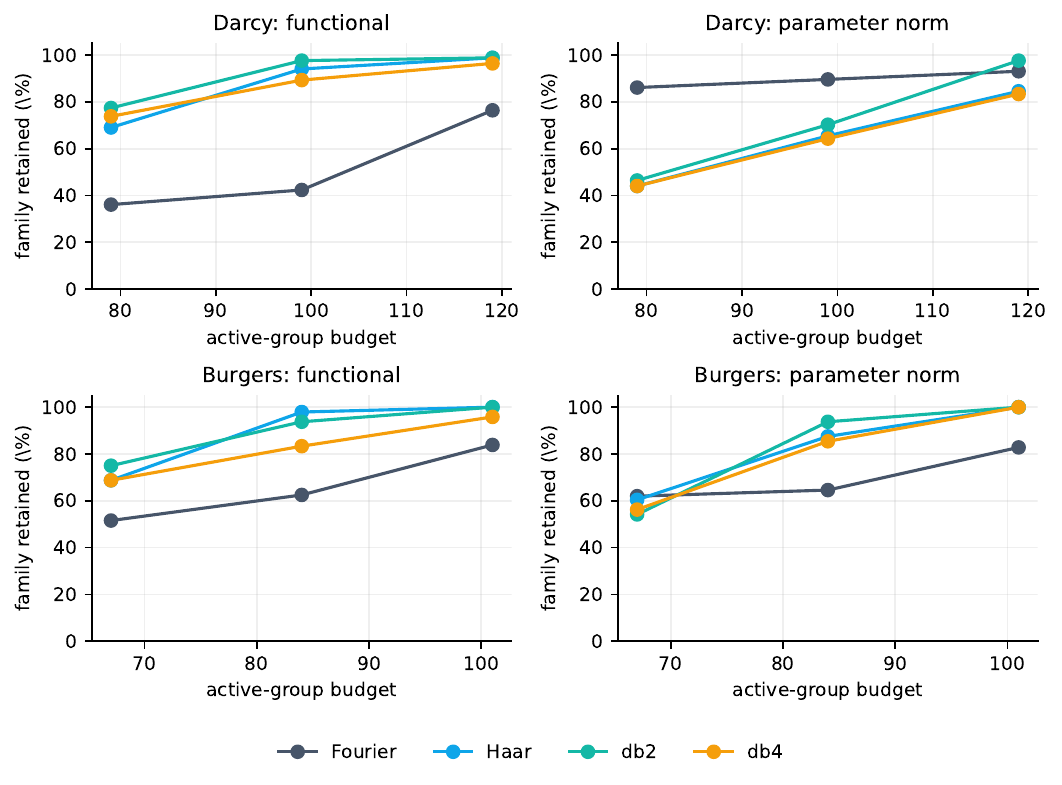}
  \caption{Distribution of retained operator families across sparsity budgets.
  Functional selection changes both the number and the type of groups retained,
  whereas parameter-norm selection produces a different allocation across
  Fourier and wavelet families.}
  \label{fig:family-budget-paths}
\end{figure}

\end{document}